\documentclass{article}

\usepackage{arxiv}

\usepackage[utf8]{inputenc} 
\usepackage[T1]{fontenc}    
\usepackage{hyperref}       
\usepackage{url}            
\usepackage{booktabs}       
\usepackage{amsfonts}       
\usepackage{nicefrac}       
\usepackage{microtype}      
\usepackage{lipsum}		
\usepackage{graphicx}
\usepackage{natbib}
\usepackage{doi}

\usepackage{multirow}
\usepackage{hyperref}
\hypersetup{
    colorlinks=true,
    linkcolor=blue,
    filecolor=magenta,      
    urlcolor=cyan,
    pdftitle={Overleaf Example},
    pdfpagemode=FullScreen,
    }

\AtBeginDocument{%
  \providecommand\BibTeX{{%
    \normalfont B\kern-0.5em{\scshape i\kern-0.25em b}\kern-0.8em\TeX}}}

\title{Evaluating Deep vs. Wide \& Deep Learners as Contextual Bandits for Personalized Email Promo Recommendations}

\author{Aleksey A. Kocherzhenko \\
	\texttt{kocherzhenko@gmail.com} \\
	\And
	Nirmal Sobha Kartha\\
	\texttt{nirmal@launchpad.ai} \\  
	\And
	Tengfei Li \\
	\texttt{davidleo1984@gmail.com}
	\And
	Hsin-Yi (Jenny) Shih \\
	\texttt{jennywho86@gmail.com}
	\And
	Marko Mandic \\
	\texttt{mm5305@columbia.edu}
	\And
	Mike Fuller \\
	\texttt{mikefullergm@gmail.com}
	\And
	Arshak Navruzyan\thanks{LaunchPad.AI, Honolulu, HI, USA} \\
	\texttt{arshak@launchpad.ai}
    }

\date{}

\begin{document}

\maketitle

\begin{abstract}
  Personalization enables businesses to learn customer preferences from past interactions and thus to target individual customers with more relevant content. We consider the problem of predicting the optimal promotional offer for a given customer out of several options
  as a contextual bandit problem. Identifying information for the customer and/or the campaign can be used to deduce unknown customer/campaign features that improve optimal offer prediction. Using a generated synthetic email promo dataset, we demonstrate similar prediction accuracies for (a) a wide and deep network that takes identifying information (or other categorical features) as input to the wide part and (b) a deep-only neural network that includes embeddings of categorical features in the input. Improvements in accuracy from including categorical features depends on the variability of the unknown numerical features for each category. We also show that selecting options using upper confidence bound or Thompson sampling, approximated via Monte Carlo dropout layers in the wide and deep models, slightly improves model performance.
\end{abstract}

\section{Introduction}
A personalized recommendation model can be conceptualized as a ranking system that learns from customers' past engagements and interactions. \cite{Pu2011}\cite{Polatidis2013RecommenderST}. Recommendation systems are a critical component of the digital marketing industry, where they are used for matching digital ads to customer preferences. In 2019, this industry was valued at 43.8 billion USD, with an expected compound annual growth rate of 17.4\% for 2020--2027 \cite{DigitalMarketing2020}.

Deep neural networks are extensively used in personalized recommendation systems. \cite{Zhang2019} However, although they are good at \textit{generalization} (prediction for previously unseen feature combinations), they struggle with \textit{memorization} (prediction based on co-occurrences of previously observed
categorical feature values) \cite{cheng2016wide}.
Wide and deep networks that use wide linear models to memorize sparse feature cross-products and deep neural networks to generalise to unseen feature values help solve this problem \cite{Zhang2019}.

Recommendation models are typically trained in an online setting, so as to learn from shifting user preferences. However, because predictions aid decision-making in the production context and influence customer decisions, they can create positive feedback loops that alter the underlying data generating process and result in a selection bias for the input data \cite{DBLP:journals/corr/abs-1710-11214}. Causal inference \cite{JMLR:v14:bottou13a} and bandit theory \cite{Li_2010} can be used to account for such algorithmic bias.

The current paper makes the following contributions:
\begin{itemize}
\itemsep0em 
\item We propose a strategy for generating synthetic email promo datasets with clearly defined measures for unknown deterministic features and feature randomness.
\item We evaluate the performance of the wide \& deep, wide-only and deep-only architectures.
\item We show that bandit algorithms that use upper confidence bound (UCB) \cite{DBLP:journals/corr/abs-1209-3352} and Thompson sampling (TS)\cite{Auer2002} for action selection slightly improves model performance.
\end{itemize}
Our code is publicly available for use by future researchers.

\section{Problem Formulation}
A business regularly runs marketing campaigns and for each campaign needs to decide which of several promotional offers might most appeal to each customer. We assume that the preferences of each customer are entirely determined by that customer’s personal characteristics, as well as those of the offers: i.e., if we knew everything about a customer and all available offers, we could predict that customer’s preferred offer with certainty. Then, there exists a generally non-linear function $f$ that deterministically maps $\mathbf{C_j}$ and $\mathbf{P_k}$, complete features of the $j^\mathrm{th}$ customer and $k^\mathrm{th}$ campaign respectively, to one of $N$ available promotional offers, $\mathbb{A} = \left\lbrace 1, ..., N\right\rbrace$:
\begin{equation}
    \label{function}
    f \left( \mathbf{C}_j, \mathbf{P}_k \right) = a \in \mathbb{A}
\end{equation}

The number of features that completely describe a customer (or a promotional campaign) is enormous, so only a \emph{subset} of features will ever be known. If $L$ features are known for each customer and $M$ features are known for each campaign, the projections of $\mathbf{C}_j$ and $\mathbf{P}_k$ onto the sub-spaces of \emph{known} customer and campaign features are an $L$-dimensional vector and an $M$-dimensional vector, respectively. We denote corresponding vectors in the full feature space, with non-zero components representing the \emph{known} features as $\mathbf{c}_j^L$ and $\mathbf{p}_k^M$. We denote vectors in the full feature space that represent the $L'$ \emph{unknown} customer features and $M'$ \emph{unknown} campaign features as $\mathbf{c}_j^{L'}$ and $\mathbf{p}_k^{M'}$, respectively. Then,

\begin{equation}
\label{vectors}
\mathbf{C}_j = \mathbf{c}_j^L + \mathbf{c}_j^{L'},\;\;\;\;\; \mathbf{P}_k = \mathbf{p}_k^M + \mathbf{p}_k^{M'}.
\end{equation}

Given unique IDs $j$ and $k$ for each customer and promotional campaign respectively, there exists a \emph{one-to-one} mapping between each ID and the corresponding complete feature vector: $j \leftrightarrow \mathbf{C}_j$, $k \leftrightarrow \mathbf{P}_k$. The existence of such mappings requires the existence of mappings of IDs to \emph{each of the components} of the corresponding vectors (but not vice versa). Therefore, there must exist functions
\begin{equation}
\label{id_functions}
g:~g\left( j \right) = \mathbf{c}_j^{L'}, \;\;\;\;\; h:~h\left( k \right) = \mathbf{p}_k^{M'},
\end{equation}
that are generally nonlinear and not necessarily invertible.

Introducing Eqs.~(\ref{vectors}--\ref{id_functions}) into Eq.~(\ref{function}), we find:
\begin{equation}
\label{function_gen}
f \left( \mathbf{C}_j, \mathbf{P}_k \right) = f \left[ \mathbf{c}_j^L + g\left( j \right), \mathbf{p}_k^M + h\left( k \right) \right] = F \left( \mathbf{c}_j^L, \mathbf{p}_k^M, j, k \right) = a,
\end{equation}
where $F$ is some new nonlinear function. In the limit when no customer or promotional campaign features are known, but it is known that all relevant features are uniquely defined by the customer ID and the campaign ID, Eq.~(\ref{function_gen}) reduces to
\begin{equation}
\label{function_gen_limit}
F \left( j, k \right) = a.
\end{equation}

Optimal offers for previously unseen customers and promotional campaigns can be learned more quickly based on similarity to previously seen customers and promotional campaigns (generalization). The better the vectors $\mathbf{c}_j^L$ and $\mathbf{p}_k^M$ approximate the vectors $\mathbf{C}_j$ and $\mathbf{P}_k$, and the better the space of all customer and promotional campaign features is sampled, the more accurately the optimal offer for any new customer-promotional campaign pair can be inferred. Observing optimal offers for \emph{specific} customer-promotional campaign pairs allows refining predictions for the rare cases when subtle differences between similar customer-promotional campaign pairs lead to different optimal offers. If similarity between different customer-promotional campaign pairs cannot be established \emph{a priori}, as is the case for Eq.~(\ref{function_gen_limit}), no generalization is possible and the optimal offer for each customer-campaign pair must be memorized individually, making the training process inefficient. With partial information about customer and promotional campaign features, memorization and generalization can be combined to enhance prediction accuracy and training efficiency.

\section{Related Work}
The idea of combining wide linear models with deep neural networks for building robust recommendation systems that can memorize as well as generalize was first proposed in Ref.~\cite{cheng2016wide}, refining the idea of factorization machines \cite{Rendle2012}. The system was deployed and evaluated on the Google Play app store, which yielded a positive app acquisition gain of 3.9\% in an online setting. 

Personalized dynamic recommendation has been reframed in a contextual bandit setting to mitigate the cold-start problem and algorithmic bias \cite{Li_2010}. This paper also developed the LinUCB algorithm to efficiently estimate confidence bounds in closed form. Because computing the actual posterior distribution of the bandit problem is usually intractable, Refs.~\cite{Dielman1994} and \cite{gal2016dropout} proposed sampling techniques that use Monte Carlo (MC) dropout to approximate the posterior distribution. Both UCB and TS on this approximated distribution perform well and are computationally efficient. Empirical evaluation of many techniques for approximating TS and the best practices for designing such approximations are discussed in Ref.~\cite{riquelme2018deep}. Recently, combining bandit algorithms with deep neural networks in an ad recommendation system on a proprietary dataset has also been explored \cite{guo2020deep}.

\section{Methodology}
\subsection{Generalizing via Embeddings}\label{embeddingsection}
Embeddings map categorical variables to vectors of pre-defined dimensionality and allow reducing dimensionality and encoding similarities in the data. Typically randomly initialized, they are then learned in a relevant supervised machine learning task. Including an $L_\mathrm{e}$-dimensional embedding vector for the user ID $j$ and an $M_\mathrm{e}$-dimensional embedding vector for the campaign ID $k$ as input to a deep neural network allows training them, over many occurrences of individual customers and promotional campaigns in the dataset.

The resulting embeddings may capture unknown customer and promotional campaign features that may be important for selecting the optimal offer $a$. In practice, a customer is represented by the \emph{concatenation} of the known customer feature vector and the customer ID embedding vector. Then, by construction, $\mathbf{c}_{j}^{L}$ and the vector $\mathbf{c}_j^{L_{\mathrm{e}}}$ that represents embedding in the full customer feature space span orthogonal sub-spaces and are linearly independent. Consequently, $\mathbf{c}_j^L +\mathbf{c}_j^{L_\mathrm{e}}$ spans an $\left( L + L_{\mathrm{e}} \right)$-dimensional subspace of the total customer feature space and can, in principle, approximate the complete feature vector $\mathbf{C}_j$ better than $\mathbf{c}_j^L$ that spans only an $L$-dimensional sub-space. Similar reasoning applies to using an embedding vector $\mathbf{p}_k^{M_\mathrm{e}}$ for campaign IDs.

If $\mathbf{c}_j^L$ and $\mathbf{p}_k^M$ are treated as fixed constraints, optimizing $\mathbf{c}_j^L + \mathbf{c}_j^{L_\mathrm{e}}$ and $\mathbf{p}_k^M + \mathbf{p}_k^{M_\mathrm{e}}$ only changes the embedding vectors. Because these vectors and the known feature vectors are linearly independent, training the embeddings does not double-count the known features, but rather augments them. Consequently, a nonlinear function
\begin{equation}
\label{approx_with_embeddings}
\widetilde{f}_\mathrm{e} \left( \mathbf{c}_j^L + \mathbf{c}_j^{L_\mathrm{e}},~\mathbf{p}_k^M + \mathbf{p}_k^{M_\mathrm{e}} \right) = \widetilde{a}_\mathrm{e} \in \mathbb{A}
\end{equation}
that is learned using both the known feature vectors and the embedding vectors should return better predictions than the nonlinear function learned only using known feature vectors.

\subsection{Memorizing Cross-Products}\label{crossproductsection}
Instead of embeddings, one can also use cross-products of customer and campaign IDs \cite{cheng2016wide} (treated as categorical features) in the wide part of a wide \& deep neural network. We must explicitly make the assumption that the optimal offer is specific to every customer-promotional campaign \emph{pair}, similarly to Eq.~(\ref{function_gen_limit}). 

If sufficient data for \emph{all} possible user ID-campaign ID pairs is available, a wide linear model is sufficient and adding a deep neural network provides no benefit for predicting the optimal offer. However, it is impossible to train the model on all possible user ID-campaign ID pairs if new customers are continuously added and new promotional campaigns are regularly run. Furthermore, if there is \emph{variability} in customer or campaign features with a given user or campaign ID, then such variability cannot captured by $F$ in Eq.~(\ref{function_gen_limit}).

In section \ref{embeddingsection}, the known customer and promotional campaign features were augmented with user ID and campaign ID embeddings that capture otherwise unknown customer and promotional campaign features, once sufficiently trained. In this approach, cross-products of user IDs and campaign IDs are input to a wide linear model, $l \left( j \times k \right)$, whereas the known customer and promotional campaign feature vectors are input to a deep nonlinear model, $\widetilde{f}_\mathrm{d} \left( \mathbf{c}_j^L, \mathbf{p}_k^M \right)$, in the same way as in Eq.~(\ref{approx_with_embeddings}). The outputs of the linear and nonlinear transformations are then passed to the output layer:
\begin{equation}
\label{deepwide}
\sigma \left[ \widetilde{f}_\mathrm{d} \left( \mathbf{c}_j^L, \mathbf{p}_k^M \right), l \left( j \times k \right) \right] = \widetilde{a}_\mathrm{dw} \in \mathbb{A},
\end{equation}
where $\sigma$ is a nonlinear activation function. 

\subsection{Generated Dataset}
The architectures discussed in Sections \ref{embeddingsection} and \ref{crossproductsection} represent different approximations of Eq.~(\ref{function_gen}). We hypothesize that, given sufficient training data, both should result in similar prediction accuracy. To test this hypothesis, we generated a synthetic dataset of 10 million email promo campaign instances with the following features for each sample: user/customer ID;
customer features 1 and 2; hidden customer feature;
campaign ID; campaign features 1 and 2; hidden campaign feature.

\begin{figure}[b]
\centering
\includegraphics[scale=0.33]{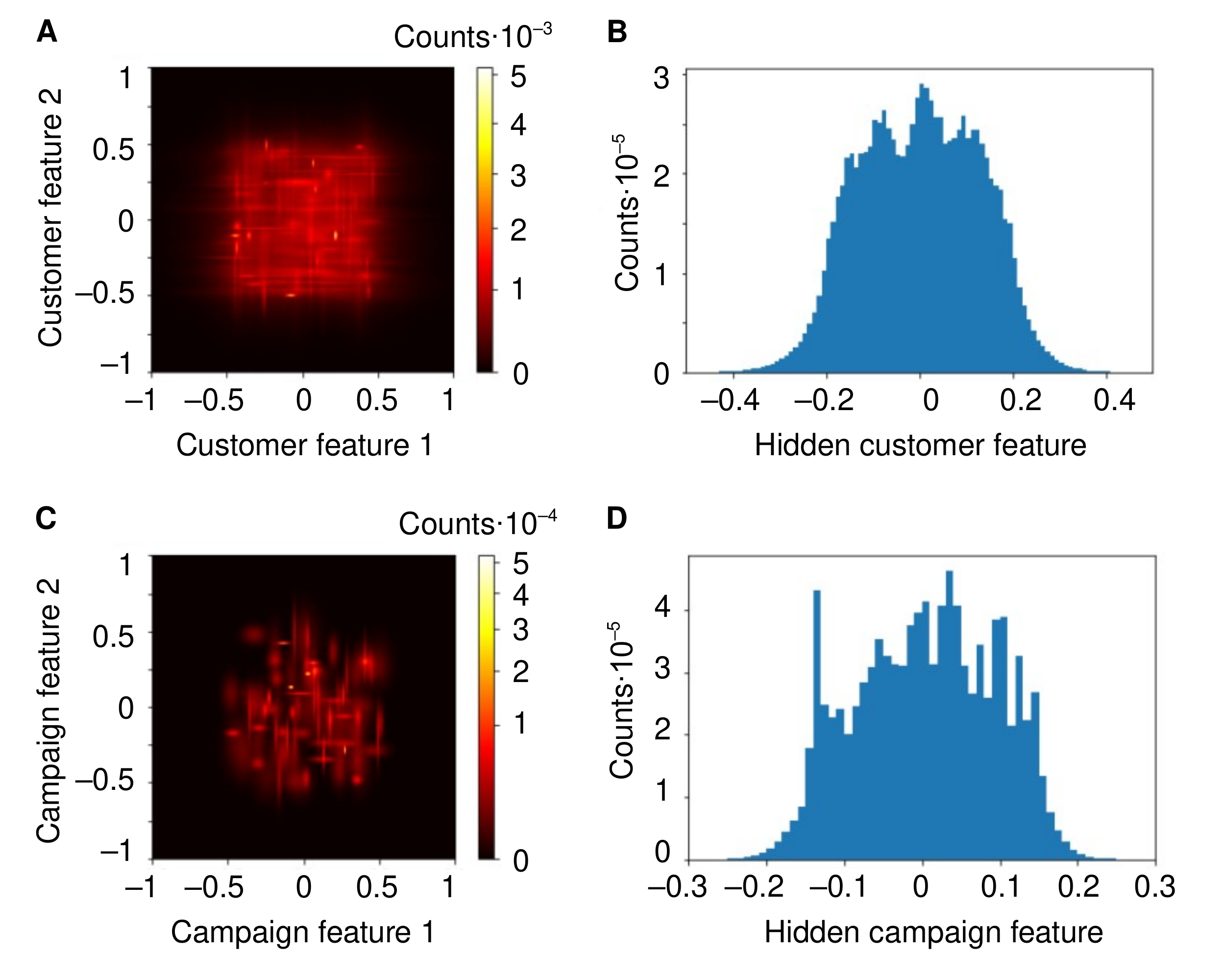}
\caption{Distributions of customer (top) and campaign (bottom) features in the dataset. Left - combinations of known feature values; right - hidden feature value.}
\label{fig_features}
\end{figure}

The dataset contains 1,000 distinct customers, each identified by a unique ID and seen a random number of times, with occurrences randomly distributed throughout the dataset. The relative frequency of seeing a specific customer is sampled from a uniform distribution on the interval (0, 1); the minimum and maximum numbers of times that any single customer is seen are 17 and 20,004, respectively. Every customer is described by three features: customer features 1 and 2, as well as a hidden customer feature. For a given customer, the mean values and the standard deviations for each of these features are constant. The mean values of customer features 1 and 2 are sampled from a uniform distribution on the interval (--0.5, 0.5). The standard deviations for these features are sampled from uniform distributions on the intervals (0, 0.2) and (0, 0.3), respectively. The mean value of the hidden customer feature is sampled from a uniform distribution on the interval (–0.2, 0.2), and the standard deviation is sampled from a uniform distribution on the interval (0, 0.1). The values of all customer features vary from sample to sample and are sampled from normal distributions with fixed means and standard deviations for each customer. The numbers of samples with each combination of customer features 1 and 2 are shown in Fig.\ref{fig_features}A; the numbers of samples with each value of the hidden customer feature are shown in Fig.\ref{fig_features}B.

The samples are evenly distributed among 100 promotional campaigns, each identified by a unique campaign ID. Each promotional campaign consists of 100,000 samples, and the promotional campaigns are listed sequentially. Every promotional campaign is described by three features: campaign features 1 and 2, and hidden campaign feature. For a given promotional campaign, the mean values and the standard deviations for each of these features are constant. The mean values for campaign feature 1 and campaign feature 2 are sampled from a uniform distribution on the interval (--0.5, 0.5). The standard deviations for these features are sampled from uniform distributions on the intervals (0, 0.1) and (0, 0.05), respectively. The mean value for the hidden campaign feature is sampled from a uniform distribution on the interval (--0.15, 0.15), and the standard deviation is sampled from a uniform distribution on the interval (0, 0.05). The values of all promotional campaign features for individual samples in the dataset are sampled in the same way as for customer features. The numbers of samples with each combination of campaign feature 1 and campaign feature 2 are shown in Fig.~\ref{fig_features}C; the numbers of samples with each value of the hidden campaign feature are shown in Fig.~\ref{fig_features}D.

As evident from Fig.~\ref{fig_features}, the customer feature space is sampled better than the campaign feature space because there are 1,000 distinct customers (compared to 100 distinct campaigns) and larger standard deviations. There is a one-to-one correspondence between the user ID and the \emph{mean values} and \emph{standard deviations} for each customer feature (likewise for the promotional campaign features). The actual feature values describing a specific occurrence of a given customer or promotional campaign are \emph{sampled} from more or less narrow normal distributions, with fixed defined means and standard deviations. This ensures variation in the customer features between different samples with the same user ID, and in the promotional campaign features between different samples with the same campaign ID. This construction reflects the fact that, in practice, the values of many customer and promotional campaign features may vary somewhat between appearances of a customer or a promotional campaign in the dataset due to unaccounted for factors that are external to the customer--promotional campaign system and reflect its interaction with the broader ``environment".

\begin{figure}[b]
\centering
\includegraphics[scale=0.3]{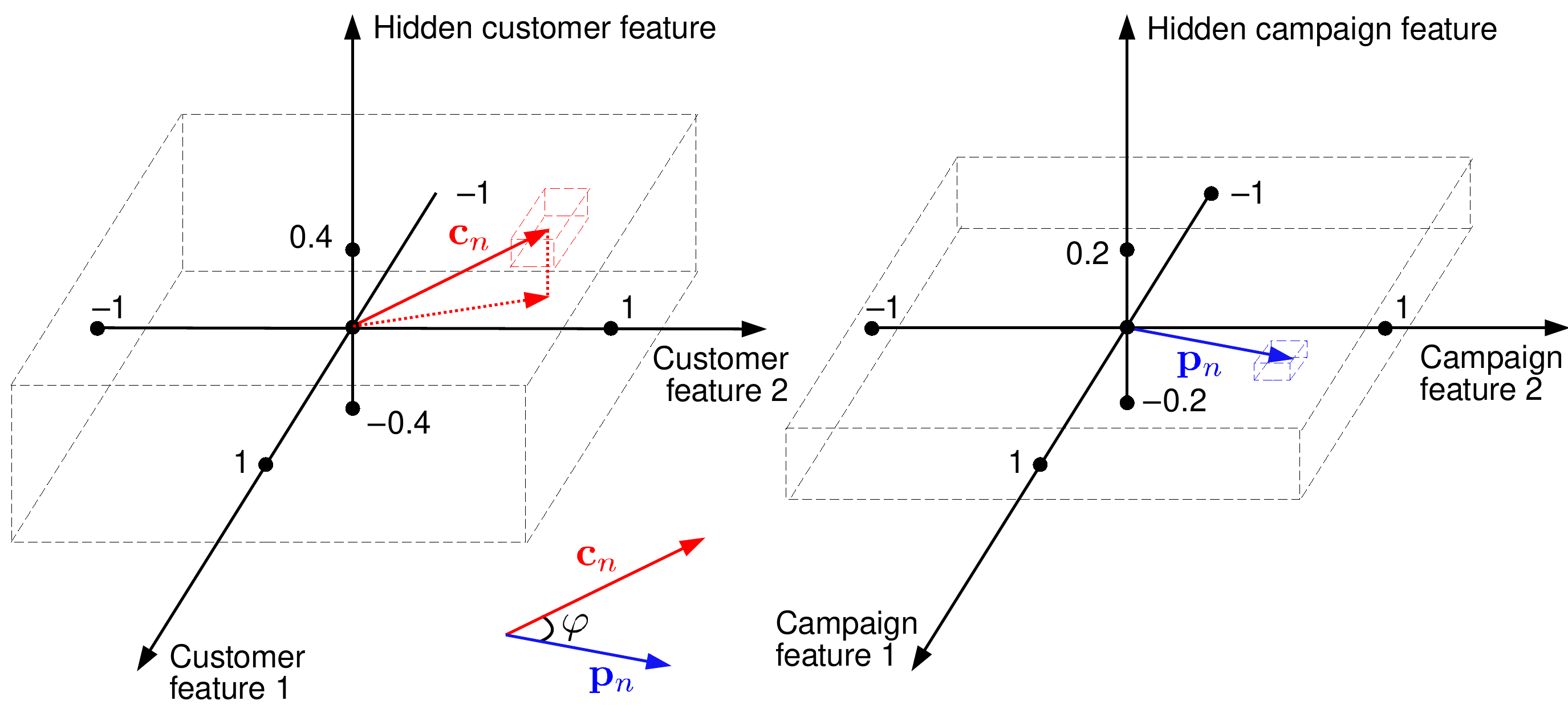}
\caption{Schematic representation of customer feature vector $\mathbf{c}_n$ (left) and campaign feature vector $\mathbf{p}_n$ (right). Dashed black boxes show the range of possible feature combinations. Dashed colored boxes illustrate the range of feature variability for a given customer or campaign. Dashed lines show projections of the complete feature vector.}
\label{fig_vectors}
\end{figure}

The three customer features for the $n$th sample in the dataset can be represented by a three-dimensional customer feature vector, $\mathbf{c}_n$, and the three promotional campaign features for the $n$th sample can be represented by a three-dimensional promotional campaign feature vector, $\mathbf{p}_n$, where the sample number $n = 1, ..., 10,000,000$ (these vectors are schematically illustrated in Fig.~\ref{fig_vectors}).

The optimal offer for the $n$-th sample is defined by
\begin{equation}
\label{opt_opt}
a_n = \left\lbrace
\begin{array}{ll}
1, & \mathrm{if}~\mathbf{c}_n \uparrow\uparrow \mathbf{p}_n, \\
\mathrm{ceil} \left[ \frac{10}{\pi} \arccos \left( \frac{\mathbf{c}_n \cdot \mathbf{p}_n}{\left| \mathbf{c}_n \right| \left| \mathbf{p}_n \right|} \right) \right], & \mathrm{otherwise},
\end{array}
\right.
\end{equation}
a piecewise continuous nonlinear function of the angle between the vectors $\mathbf{c}_n$ and $\mathbf{p}_n$ that can take 10 values, $a_n \in \left\lbrace 1, ..., 10\right\rbrace$. For two samples $n$ and $n'$, where the customers and promotional campaigns have similar features, the optimal offers defined by Eq.~(\ref{opt_opt}) will be the same, $a_n = a_{n'}$, unless the angles between their customer and promotional campaign feature vectors happen to be separated by a point of discontinuity (in which case $\left| a_n - a_{n'} \right| = 1$).

All models in our experiment were trained on 60\% of the generated dataset, with 20\% used for validation and 20\% used for testing; the data was split into the training, validation, and test sets \emph{randomly}. For comparison, a different split of the dataset was also used for training and testing two of the models: the \emph{first} 80\% of the data was randomly split into the training and validation datasets at a 3:1 ratio, and the \emph{remaining} 20\% was used for testing.

\subsection{Model Architectures}\label{modelarchitectures}
This paper focuses on 3 network architectures: a wide model, a deep model, and a wide \& deep model.

\begin{figure}
\centering
\includegraphics[scale=0.3]{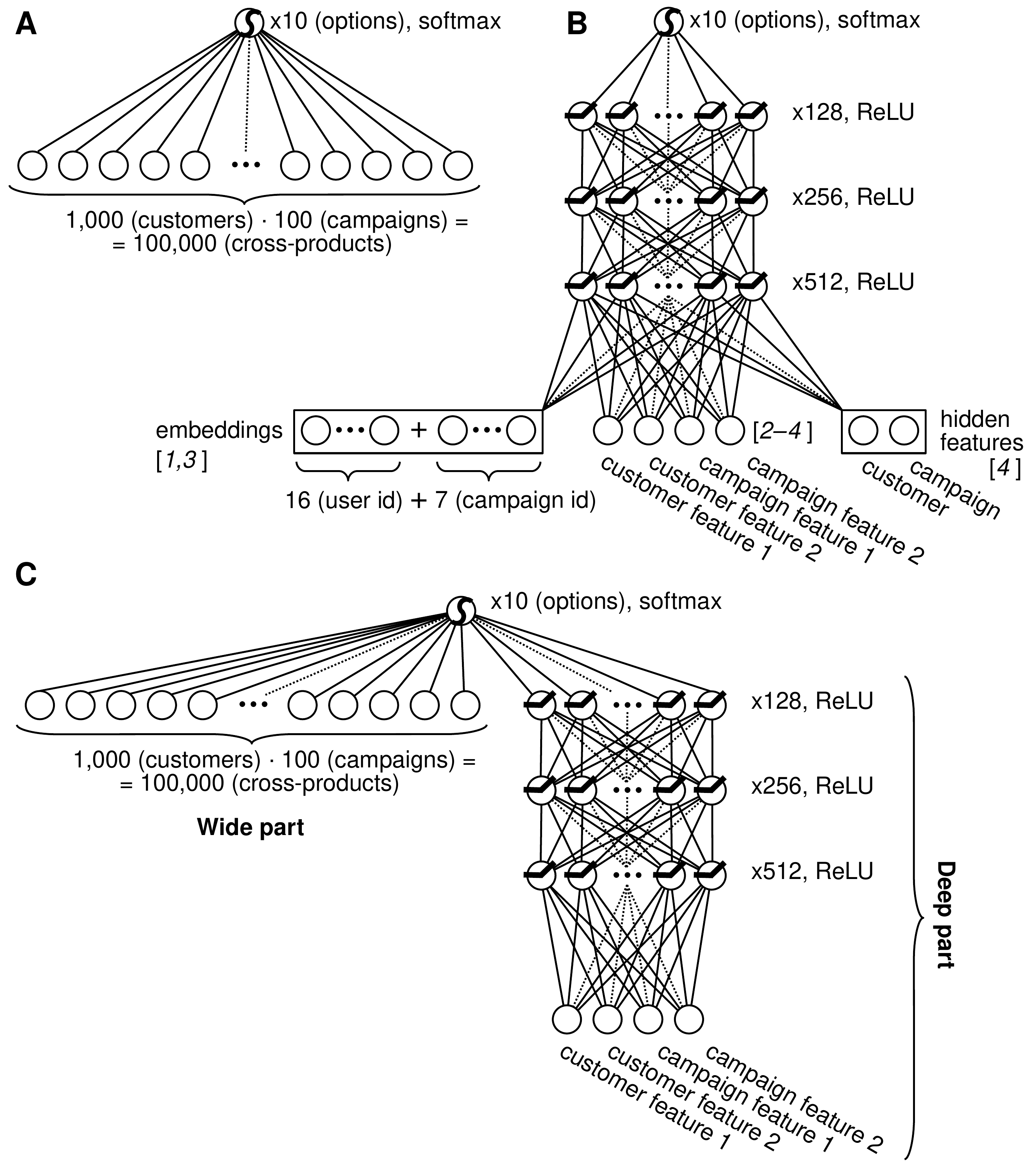}
\caption{Architectures used for the wide (A), deep (B) and wide \& deep models (C). Italic numbers in square brackets indicate model versions with different inputs (See Table~\ref{tab_acc}).}
\label{fig_models}
\end{figure}

The wide model (Fig.~\ref{fig_models}A) took as input 100,000 cross-products between the 100 distinct user IDs and the 1,000 distinct campaign IDs in the dataset. This input was passed to a single linear transformation and a softmax function to select one of the ten possible offers. This model had 1 million trainable parameters.

The deep model (Fig.~\ref{fig_models}B) used a dense neural network with three hidden layers (with 512, 256, and 128 nodes and ReLU activation functions), followed by a softmax output layer to select among the ten possible offers. This model was trained with four different sets of features in the input:

\begin{enumerate}
\itemsep0em 
\item 16-dimensional embeddings for user IDs and 7-dimensional embeddings for campaign IDs;
\item customer features 1 and 2 and campaign features 1 and 2;
\item customer features 1 and 2, campaign features 1 and 2, as well as 16-dimensional embeddings for user IDs and 7-dimensional embeddings for campaign IDs;
\item customer features 1 and 2, campaign features 1 and 2, as well as the hidden customer and campaign features.
\end{enumerate}

Even with the largest input dimensions (version 3), this model had under 200 thousand trainable parameters (considerably fewer than the wide model). We found that increasing or decreasing the number of nodes in all hidden layers by a factor of 2 had no statistically significant effect on the resulting prediction accuracy.

For the wide and deep model (Fig.~\ref{fig_models}C), the cross-products between user IDs and campaign IDs (wide input) were first concatenated with the output of the third hidden layer of the deep model and then passed to the output softmax layer. Only customer features 1 \% 2 and campaign features 1 \& 2 were used as input for the deep model.

TensorFlow \cite{tensorflow2015-whitepaper} implementations of these models were trained using categorical cross-entropy loss and the Adam optimizer with a learning rate of $10^{-3}$. Early stopping with a patience of 5 epochs (deep, wide) or 10 epochs (wide \& deep) was applied. Varying the patience between 5 and 20 epochs had no significant effect on the resulting accuracy. Early stopping occurred after 16 epochs for the wide model, after 46, 33, 41, and 14 epochs respectively, for the four sets of inputs to the deep model, and after 108 epochs for the wide \& deep model.

\subsection{Bandit Implementation}\label{banditbuild}
A contextual bandits predicts an action that should be taken for a specific context in order to maximize a reward. In selecting an action, the bandit algorithm balances \emph{exploitation} of prior knowledge about the rewards associated with various possible actions and \emph{exploration} of actions that have not been tried enough times to confidently predict their associated rewards.

In the scenario considered in this paper, contextual bandits are applied to a multi-class classification problem, where the objective is to find the mapping of the customer and campaign features that serve as the context onto the optimal action, in accordance with Eq. (\ref{function_gen}). The probabilities of each action being optimal are returned by the softmax output layer of a neural network with one of the architectures described in Section~\ref{modelarchitectures}. We treat these probabilities as the normalized rewards associated with each action.

We consider two popular bandit algorithms, Thompson sampling (TS) and upper confidence bound (UCB), and compare their performance to that of a neural network that does not use a bandit algorithm. The TS algorithm randomly samples a single value from the posterior distribution of the rewards associated with each action, then selects the action with the highest sampled reward. The UCB algorithm optimistically selects the action with the highest upper confidence bound for its reward value. The confidence bounds depend on the expected rewards associated with each action and on their uncertainties that can be estimated from the posterior distributions of rewards.

The posterior distributions of rewards are approximated using the Monte Carlo (MC) dropout method \cite{gal2016dropout}. The reward distributions are approximated by making multiple forward passes through a neural network with MC dropout at inference time. To improve computational inefficiency, MC dropout (with dropout probability $p=0.3$) is only applied in a ``multi-head'' layer that immediately precedes the softmax output layer \cite{guo2020deep}. Thus, only one forward pass through the computation graph up to the dropout layer is necessary, and multiple passes from the dropout layer onward can be executed in parallel.

To select an action using the TS algorithm, we perform a single pass through the network with MC dropout and pick the action with the largest sampled value. In the UCB algorithm at 95\% confidence level, we perform 100 forward passes starting at the MC dropout layer and approximate the upper confidence bound for each action by the $5^\mathrm{th}$-largest value among 100 samples.

\section{Results}
Table \ref{tab_acc} shows the accuracy of the optimal offer predictions for all models, trained and tested on random subsets of our dataset.

\begin{table*}
  \centering
  \caption{Performance Comparison: Wide vs Deep vs Wide \& Deep architectures}
  \label{tab_acc}
  \begin{tabular}{p{20mm}p{48mm}p{35mm}p{10mm}p{10mm}p{10mm}}
    \toprule
    \multirow{2}{*}{\textbf{Model}} & \multirow{2}{*}{\textbf{Deep input}} & \multirow{2}{*}{\textbf{Wide input}} & \multicolumn{3}{c}{\textbf{Accuracy, \%}} \\
    \cline{4-6}
    & & & \textbf{Train} & \textbf{Valid} & \textbf{Test} \\
    \midrule
    \rule{0pt}{1ex}    
    {Random} & None & None & 10.0 & 10.0 & 10.0\\
    \cline{1-6}
    \rule{0pt}{3ex}    
    {Wide} & None & user ID $\times$ campaign ID & 35.2 & 33.3 & 33.3 \\
    \cline{1-6}
    \rule{0pt}{3ex}    
    \multirow{4}{*}{Deep} & $^1$User \& campaign ID embeddings & \multirow{4}{*}{None} & 27.6 & 27.0 & 26.9 \\
    \cline{2-2}
    \cline{4-6}
    \rule{0pt}{3ex} 
    & $^2$Customer features 1\&2, \newline Campaign features 1\&2 &  & 68.0 & 68.1 & 68.0\\
    \cline{2-2}
    \cline{4-6}
    \rule{0pt}{3ex} 
    & $^3$Customer features 1\&2, \newline Campaign features 1\&2, \newline User \& campaign ID embeddings &  & 81.3 & 80.9 & 80.8\\
    \cline{2-2}
    \cline{4-6}
    \rule{0pt}{3ex} 
    & $^4$Customer features 1\&2, \newline Campaign features 1\&2, \newline Hidden User \& Campaign features &  & 97.7 & 98.0 & 97.2\\
    \cline{1-6}
    \rule{0pt}{3ex} 
    Wide \& Deep & Customer features 1\&2, \newline Campaign features 1\&2 & user ID $\times$ campaign ID & 80.2 & 77.3 & 77.3\\
    \bottomrule
  \end{tabular}
\end{table*}

As expected, random guessing results in 10\% accuracy for predicting the optimal offer out of 10 possible offers. Also as expected, the prediction accuracy for the deep model that takes the complete customer and promotional campaign feature vectors as input approaches 100\% (it is 97.2\% for the test set and could potentially be further improved by providing additional training data, using a larger neural network, adjusting the learning rate, etc.).  

Note that for all subsequent models, hidden customer and campaign features are \emph{not used}: only customer features 1 and 2, campaign features 1 and 2, and/or the user and campaign IDs are used for predicting the optimal offer. Thus, the models were deliberately provided with incomplete information:

\begin{itemize}
\item Numerical customer and campaign features supplied  projections of the complete customer and campaign feature vectors onto the ``known" customer and campaign feature subspaces.
\item In accordance with Eq.~(\ref{id_functions}), the user and campaign IDs can be mapped to the \emph{mean value} and \emph{standard deviation} that define the distribution of the hidden customer and campaign features for a specific customer and a specific promotional campaign. However, the user and campaign IDs \emph{cannot} be deterministically mapped to the values of the hidden customer and campaign features \emph{for a specific sample}, because these features are \emph{randomly sampled} from a normal distribution with a fixed mean and a fixed standard deviation.
\end{itemize}
Therefore, even with the user and campaign IDs available, the information that is provided to the models is incomplete.

If only identifying information for the customer (user ID) and for the promotional campaign (campaign ID) were known, any inference of the optimal offer would be based on memorization. The wide-only model with the user ID and campaign ID cross products as input correctly predicts the optimal offer in about a third of cases, while the deep-only model with only the user ID and campaign ID embeddings as input correctly predicts the optimal offer in just over a quarter of all cases (see Table~\ref{tab_acc}). The differences in accuracy between the two models are not due to difference in the number of trainable parameters, as varying layer sizes for the deep model did not significantly affect prediction accuracy. These differences are also not due to insufficient training data, as training both models on smaller subsets did not significantly change accuracies either. Therefore, the wide model outperforms the deep model when it comes to pure memorization, at least for the embedding sizes used for the user ID and the campaign ID. This difference in accuracy may be due to more efficient memorization for the wide model, compared to a deep model, where the user ID and the campaign ID embeddings are randomly initialized and then adjusted every time a user ID or campaign ID value is encountered in the training data.

Note that the poor performance of both the wide and the deep models trained using only the user and campaign IDs is largely determined by the relatively broad distributions of the customer (campaign) features for a given user ID (campaign ID). For instance, reducing the maximum standard deviations for customer feature 1, customer feature 2, and the hidden customer feature from 0.2, 0.3, and 0.1 respectively, to 0.05, 0.1, and 0.05 respectively, without changing the campaign feature distributions, improves the performance of the deep model from 26.9\% to 62.1\%. If the customer and campaign features were completely determined by the user and campaign IDs, in accordance with Eq.~(\ref{id_functions}), then near-perfect prediction accuracy could, in principle, be achievable for both the wide and deep models, provided that they were sufficiently trained.

\begin{table*}[h]
\centering
  \caption{Performance Comparison: Different Feature Distributions for Training/Validation and Test Sets}
  \label{tab_acc_seq}
  \begin{tabular}{p{20mm}p{48mm}p{35mm}p{10mm}p{10mm}p{10mm}}
    \toprule
    \multirow{2}{*}{\textbf{Model}} & \multirow{2}{*}{\textbf{Deep input}} & \multirow{2}{*}{\textbf{Wide input}} & \multicolumn{3}{c}{\textbf{Accuracy, \%}} \\
    & & & \textbf{Train} & \textbf{Valid} & \textbf{Test} \\
    \midrule
    \rule{0pt}{1ex} 
    Deep & Customer features 1\&2, \newline Campaign features 1\&2, \newline User \& campaign ID embeddings & None & 81.9 & 81.2 & 58.1 \\
    \cline{1-6}
    \rule{0pt}{3ex} 
    Wide \& Deep & Customer features 1\&2, \newline Campaign features 1\&2 & user ID $\times$ campaign ID & 80.0 & 77.6 & 63.2 \\
    \bottomrule
  \end{tabular}
\end{table*}

If only the projections of the complete customer and campaign feature vectors onto the known customer and campaign feature subspaces (customer features 1 and 2, campaign features 1 and 2) were known, then the accuracy of the optimal offer predictions would largely be determined by how well these projections approximate the complete feature vectors. In the limit of the hidden customer and hidden campaign features being zero for all samples, customer features 1 and 2 and campaign features 1 and 2 would provide complete information about the customer and the promotional campaign. Consequently, near-perfect accuracy for this model could be expected, for a sufficiently large and sufficiently trained neural network. When the deep model only takes as input customer features 1 and 2 and campaign features 1 and 2 in our generated dataset that has relatively large hidden customer and campaign features, it can predict the optimal offer in over two thirds of cases (see Table~\ref{tab_acc}). Further improving the prediction accuracy requires accounting for the hidden customer and campaign features that can, to an extent, be inferred from the user and campaign IDs. The precision of this inference is limited by the variability in the hidden features for a given customer (promotional campaign).

Predictably, the highest accuracy of the optimal offer predictions were achieved when both the identifying information for the customer and the promotional campaign, as well as the projections of the complete customer and campaign feature vectors onto the known customer and campaign feature subspaces were provided to the machine learning models. The deep model that took customer features 1 and 2, campaign features 1 and 2, as well as the user ID and campaign ID embeddings as input achieved an accuracy of 80.8\% on the test set. The wide \& deep model that took customer features 1 and 2 and campaign features 1 and 2 as input to the deep part, as well as user ID--campaign ID cross-products as input to the wide part, achieved an accuracy of 77.3\% on the test set. Interestingly, the deep-only model slightly outperformed the wide \& deep model, even though the wide model that took user ID--campaign ID cross-products as input outperformed the deep model that took only user ID and campaign ID embeddings as input. Examining the confusion matrices for the predicted optimal offer, $\widetilde{a}\in\mathbb{A}$, vs. the actual optimal offer, $a\in\mathbb{A}$, suggests that the performance of both models is largely limited by the randomness in the hidden customer and campaign features: $\left| a - \widetilde{a} \right| \le 1$ for over 95\% of the predictions.

It should be noted that the reasonably high accuracies for the deep-only and the wide \& deep models that took customer features 1 and 2, campaign features 1 and 2, as well as the user and campaign IDs as input were achieved because the training, validation, and test sets were \emph{randomly sampled} from the generated dataset. Consequently, the training, validation, and test sets all contained samples with (nearly) all possible user ID--campaign ID combinations. Randomly splitting the \emph{first} 80\% of the data between the training and validation sets at a 3:1 ratio and using the \emph{remaining} 20\% as the test set leads to significantly worse prediction accuracies on the test set for both the deep-only and the wide \& deep models (Table~\ref{tab_acc_seq}). In this case, the model is trained and tested on \emph{different} identifying information (different campaigns), making memorization useless, and, in fact, counterproductive: the accuracies on the test set for both the deep-only and the wide \& deep models turn out to be considerably lower than the accuracy for the deep model that just uses customer features 1 and 2 and campaign features 1 and 2 as input (Table~\ref{tab_acc}). In practice, this result means that considering a cross-product of a user ID with a campaign \emph{type}, for which certain features are similar across multiple campaigns, is likely to lead to enhanced performance of a machine learning model. Conversely, using a cross-product of a user ID with a campain ID that will not re-occur in the data to which the model will eventually be applied may be detrimental to the model's prediction accuracy.

We evaluated the benefit of using the implementations of UCB and TS, outlined in Section \ref{banditbuild}, on the output of the wide \& deep model for our dataset (see Table \ref{table_acc_bandit}). Both bandit algorithms show a slight improvement in accuracy compared to the wide \& deep model without explicit exploration. Our implemention of UCB consistently quantified uncertainty and predicted optimal offers better than TS, but was considerably more computationally expensive with, on average, an order of magnitude longer runtimes.

\begin{table}[t]
\centering
  \caption{Performance Comparison: Bandit Algorithms}
  \label{table_acc_bandit}
  \begin{tabular}{cccc}
    \toprule
    \multirow{2}{*}{\textbf{Model}} & \multicolumn{3}{c}{\textbf{Accuracy, \%}} \\
    & \textbf{Train} & \textbf{Valid} & \textbf{Test} \\
    \midrule
    No Bandit & 80.2 & 77.3 & 77.3 \\
    TS & 81.1 & 80.3 & 79.3 \\
    UCB & 81.6 & 80.8 & 80.2 \\
    \bottomrule
\end{tabular}
\end{table}

\section{Conclusions and Future Work}
Identifying features, such as user or campaign IDs, can significantly improve predictions of machine learning models, especially when properties of the customer and of the promotional campaign are uniquely determined by their identities. However, variability in features that correspond to the same ID reduces the predictive power of identifying features (or other categorical features that characterize the properties of individual customers or campaigns).

If the optimal offer is, to some extent, determined by the cross-product of two or more categorical features, then any improvements in the accuracy from using these features in a model hinges on the following requirement: the data used to train the model should contain the same \emph{combinations} of the categorical feature values as the test set. Otherwise, using cross-products of categorical features may be detrimental to the model's accuracy on the test set. 

Cross-products of relevant categorical features can be provided as input to the wide part of a wide \& deep neural network. Alternatively, embeddings for individual categorical features can be included in the input for a deep neural network. In the setting we explored, both models showed similar prediction accuracies and computational costs. These results may be dataset dependent, so it would be of interest to study how the accuracies and computational costs for both models depend on the total numbers of numerical and categorical features, the variability between features for a given customer and/or promotional campaign, the number of offers per campaign, the embedding sizes for categorical features, etc.

A wide \& deep model requires more careful feature engineering than a deep model with embeddings for categorical features. Specifically, it requires explicitly identifying the cross-products of categorical features that affect the optimal offer. Conversely, when using embeddings in a deep model, any interplay between categorical features is implicit in the trained network parameters. If the relevant cross-products are easily identifiable, using a wide \& deep model may be the better option. Otherwise, particularly for datasets with many categorical features, a deep model with embeddings may require less extensive exploratory data analysis.

Our network approximations of the TS and UCB bandit algorithms conferred an increase in accuracy of 1--3\% for predicting the optimal offer. However, the computational cost of bandit algorithms may complicate their adoption in online settings. Because in online settings a sound exploration strategy can limit algorithmic bias, we intend to explore possible solutions in future research.

\bibliographystyle{ACM-Reference-Format}
\bibliography{sample-sigconf}

\section*{Appendix: Online Resources}
The code used in this paper is publicly available at \newline https://github.com/fellowship/deep-and-wide-bandit.

The dataset used can be found at https://www.kaggle.com/ \newline alekseykocherzhenko/deep-with-embeddings-vs-deep-and-wide.
\end{document}